\documentclass{article} % For LaTeX2e
\usepackage{iclr2019_conference,times}

\usepackage{hyperref}
\usepackage{url}

\usepackage{amsmath,amsfonts,amssymb}
\usepackage{graphicx}
\usepackage{setspace}
\usepackage{color}%Nguyen
\usepackage{minibox}%Nguyen
\usepackage{xcolor,colortbl}%Nguyen
\usepackage{multirow}% http://ctan.org/pkg/multirow
\usepackage{hhline}% http://ctan.org/pkg/hhline
\usepackage{booktabs} %\toprule

\usepackage[]{algorithm2e}
\usepackage[utf8]{inputenc}

\usepackage[numbers]{natbib}

\newcommand\blfootnote[1]{%
  \begingroup
  \renewcommand\thefootnote{}\footnote{#1}%
  \addtocounter{footnote}{-1}%
  \endgroup
}

\title{Matching-based Depth Camera and Mirrors for 3D Reconstruction}

\author{Trong-Nguyen Nguyen\\
DIRO, University of Montreal\\
Montreal, QC, Canada\\
\texttt{nguyetn@iro.umontreal.ca}\\
\And
Huu-Hung Huynh\\
University of Science and Technology\\
Danang, Vietnam\\
\texttt{hhhung@dut.udn.vn}\\
\And
Jean Meunier\\
DIRO, University of Montreal\\
Montreal, QC, Canada\\
\texttt{meunier@iro.umontreal.ca}\\
}

\iclrfinalcopy % Uncomment for camera-ready version, but NOT for submission.
\begin{document}

\maketitle

\begin{abstract}
Reconstructing 3D object models is playing an important role in many applications in the field of computer vision. Instead of employing a collection of cameras and/or sensors as in many studies, this paper proposes a simple way to build a cheaper system for 3D reconstruction using only one depth camera and 2 or more mirrors. Each mirror is equivalently considered as a depth camera at another viewpoint. Since all scene data are provided by only one depth sensor, our approach can be applied to moving objects and does not require any synchronization protocol as with a set of cameras. Some experiments were performed on easy-to-evaluate objects to confirm the reconstruction accuracy of our proposed system.\blfootnote{Trong-Nguyen Nguyen, Huu-Hung Huynh, and Jean Meunier, ``Matching-based depth camera and mirrors for 3D reconstruction'', Proc. SPIE 10666, Three-Dimensional Imaging, Visualization, and Display 2018, 1066610 (16 May 2018); https://doi.org/10.1117/12.2304427}
\end{abstract}

\section{Introduction}\label{sec:kinect1intro}

Compared with 2D image, processing 3D information usually requires more computations as well as more resources such as memory and storage capacity. With the strong development of electronic devices in term of processing speed, many vision-based applications are now focusing on 3D data in order to exploit more information. Some researchers performed the reconstruction based on a sequence of images captured by a camera at different positions \cite{Pollefeys2000}. An obvious drawback of such methods is that the object of interest has to be static. Therefore in order to deal with moving objects, many recent studies employed a system of multiple color cameras \cite{Khan2007} and/or depth sensors \cite{Auvinet2012}. The main disadvantage of such approaches is that they require a synchronization protocol (e.g. Refs.~\citenum{Auvinet2012} and~\citenum{Duckworth2011}) when working on moving objects, and sometimes each camera and/or sensor has to be connected to a unique computer. The latency of system as well as cost of devices are thus increased. In order to overcome these problems, our approach employs only one depth camera together with 2 or more mirrors for building a system for 3D reconstruction. Synchronization is not necessary since all captured data are provided by only a single device, and the complexity and cost of equipments can thus be decreased.

\begin{figure}[h]
	\centering
	\begin{picture}(250,158)
	\put(10,0){\includegraphics[scale=0.5]{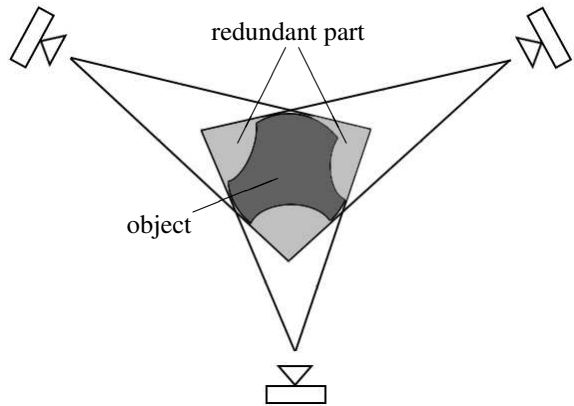}}
	\put(55,67){object}
	\put(80,75){{\line(5,2){32}}}
	\put(87,140){redundant part}
	\put(115,137){{\line(-1,-2){18}}}
	\put(120,137){{\line(1,-2){18}}}
	\end{picture}
	\caption{Redundancy when reconstructing a 3D object using shape-from-silhouette or space carving techniques in which the inputs are three color images. The overall gray region is the reconstruction result.}
	\label{fig:kinect1redundancy}
\end{figure}

As mentioned, a 3D reconstruction could be performed with a system of basic color cameras (e.g. convex hull). However, there are some advantages for using a depth sensor in this work. The most important one is that a depth map could indicate details on the object surface such as concave regions while a combination of object silhouettes provides a convex hull with redundancies (see Fig.~\ref{fig:kinect1redundancy}). Another reason is that our approach requires mirror calibrations, i.e. estimating mirror planes, a depth sensor thus reduces the complexity of this stage.

\begin{figure}[h]%[!htb]%[t]
	\centering
	\begin{picture}(240,235)
	\put(33,0){\includegraphics[scale=0.8]{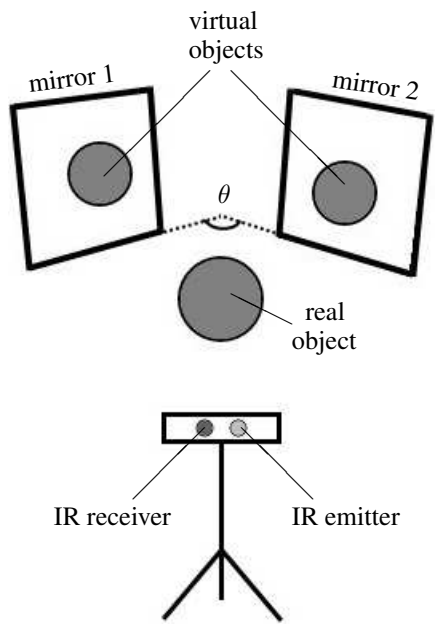}}%(70,110)
	\put(40,202){\rotatebox{7}{mirror 1}} \put(155,204){\rotatebox{-9}{mirror 2}}
	\put(140,110){\minibox{\hspace{0.18cm}real\\object}}\put(140,114){\line(-2,1){23}}
	\put(101,220){\minibox{virtual\\objects}}\put(109,210){\line(-1,-1){40}}\put(117,210){\line(1,-1){43}}
	\put(140,38){IR emitter}
	\put(148,48){\line(-1,1){27}}
	\put(50,38){IR receiver}
	\put(80,48){\line(1,1){27}}
	\put(112,160){$\theta$}
	\end{picture}
	\caption{An overview of our setup including a camera with structured-light depth estimation, two mirrors, and a sphere. The notation $\theta$ indicates the angle between the two mirror's surfaces.}
	\label{fig:kinect1scene}
\end{figure}

Depth cameras which are popularly used in vision applications could be categorized into two types: matching-based, e.g. stereo and structured-light (SL), and time-of-flight (ToF). Let us introduce briefly these two depth estimation mechanisms to explain why a depth sensor using the former technique is preferred in our approach. A matching-based approach generates a depth map by matching input images. A stereo camera captures two color images of a scene at different viewpoints while a SL-based device projects a template of light and then matches it with the corresponding image captured with a camera. Since this mechanism is related to the human vision system, we can also expect a good depth estimation of the object behind mirrors. A ToF camera uses infrared (IR) emitter and receiver to measure depth of scene based on the traveled time of a high-speed pulse or the phase shift of continuous wave. Both measurements depend on traveled trajectories of IR signals which are more difficult to predict with high-reflection surfaces such as mirror. The depth of reflected objects could thus become significantly deformed. In summary, the depth map provided by a matching-based depth camera is easier to manage than a ToF sensor in our configuration. An illustration of our setup is presented in Fig.~\ref{fig:kinect1scene}.

The remaining of this paper is organized as follows. The reliability of depth map measured by SL matching with mirrors is analyzed in Sec.~\ref{sec:kinect1reliability}. Section~\ref{sec:kinect1calibration} mentions the way of calibrating planes of mirror surfaces. Reconstructing object point cloud from depth map is presented in Sec.~\ref{sec:kinect1reconstruction}. Our experiments and evaluation are shown in Sec.~\ref{sec:kinect1testing}, and Sec.~\ref{sec:kinect1conclusion} presents the conclusion.

\section{Reliability of structured-light matching with mirrors}\label{sec:kinect1reliability}

According to geometrical optics, image of an object is reversed when seen in a mirror. As mentioned in Sec.~\ref{sec:kinect1intro}, matching-based depth measurement usually employs a passive approach such as stereo or an active one such as SL. Although stereo images of an object-behind-mirror are reversed, the matching process can be expected to provide a reliable depth map. However, we can guess that a SL-based camera may give an ambiguity since there are reflected regions in the captured image while the corresponding light pattern is still unchanged. Fortunately, this ambiguity does not happen as explained below.
\begin{figure}[h]%[!htb]%[t]
	\centering
	\begin{picture}(260,215)
	\put(50,0){\includegraphics[scale=0.5]{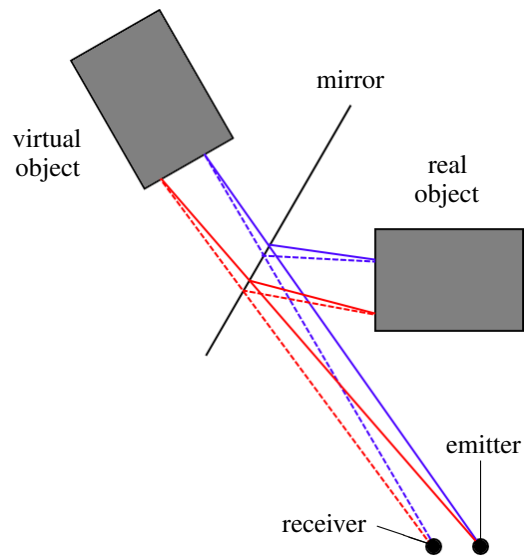}}%(70,110)
	\put(145,178){mirror}
	\put(182,140){\minibox{\hspace{0.18cm}real\\object}}
	\put(30,150){\minibox{virtual\\\hspace{0.04cm}object}}
	\put(194,40){emitter}
	\put(207,37){\line(0,-1){32}}
	\put(132,10){receiver}
	\put(168,12){\line(3,-1){22}}
	%\put(10,130){region inside mirror}
	%\put(-5,5){\includegraphics[scale=0.4]{kinect1/IRmirror.png}}
	%\put(50,127){\color{blue}\linethickness{0.4mm}\line(0,-1){50}}
	%\put(128,5){\includegraphics[scale=0.4]{kinect1/IR.png}}
	\end{picture}
	\caption{Example of emitting and receiving a structured-light pattern in a mirror. Emitter (or projector) is the source which emits the light pattern, and receiver captures the illuminated scene. The received pattern is not reversed because the rays are reflected twice.}
	\label{fig:kinect1reflect}
\end{figure}
Let us consider a configuration example in Fig.~\ref{fig:kinect1reflect}, in which the light pattern is characterized by the order of two different colored lines emitted from the projector. It is obvious to see that this pattern is flipped on the real object surface. This swap happens after the light rays touch the mirror surface. The received pattern, however, is similar to the original one (in term of order) since it touches the mirror surface twice when traveling from the emitter to the receiver. Therefore the matching result with a SL-based camera will be unaffected and reliable.%A realistic example is also presented in Fig.~\ref{fig:kinect1reflect} to show that there is almost no discontinuity in the captured IR pattern between in-mirror regions and the scene around.

\section{Mirror calibration}\label{sec:kinect1calibration}

Calibration is considered as the primary step in most vision-based applications. When dealing with a system of cameras, researchers typically perform the calibration for estimating not only the internal camera parameters but also external relationship between these cameras \cite{Auvinet2012}. Even when working on a configuration which is similar to ours, researchers also consider it as a collection of a realistic and virtual cameras \cite{Akay2014}. The proposed solutions in such studies thus employ external calibrations. Our work avoids this redundancy by estimating only internal camera matrix together with mirror surfaces based on captured depth data. The idea of using mirror planes is quite simple: object parts which are seen in a mirror will be reconstructed by reflecting them through this mirror. Since the camera calibration process has been dealt with by many approaches for color input \cite{Hartley2003} and monochromatic depth images (e.g. Ref.~\citenum{Auvinet2012}), this section only mentions the latter problem.

There are many ways for estimating the mirror plane based on the depth map provided by the camera. An indirect method could be employed by putting one or some easy-to-locate calibration objects (e.g. simple marker, cylinder, cube) in front of the mirror. The plane is then determined based on 3D coordinates of these real objects together with corresponding virtual ones behind the mirror. Another method, that directly estimates the mirror surface, is also possible. The depth map of the mirror's frame could be used to assess the position of the plane if it is large enough. In our setup, the frame is too small, thus the plane equation of each mirror surface was estimated based on 3D coordinates of some markers placed on it. Since the depth $Z$ of a pixel $(x, y)$ is given by the depth image, the corresponding point $(X,Y,Z)$ in 3D space can be localized using internal parameters of the depth camera as
\begin{equation}
	 [X,Y,Z]^\top = Z\cdot diag(f_x^{-1},f_y^{-1},1)[x-c_x,y-c_y,1]^\top
	\label{eq:kinect1reproject}
\end{equation}
where $(c_x,c_y)$ is the principal point on the image, $f_x$ and $f_y$ are focal lengths. These values can be easily estimated based on standard camera calibration techniques \cite{Szeliski2010}.

Given a set of $n$ markers, the mirror plane, which is characterized by a collection of 4 parameters $(a,b,c,d)$, is determined by solving the equation
\begin{equation}
	\begin{bmatrix}X_1 & Y_1 & Z_1 & 1\\X_2 & Y_2 & Z_2 & 1\\ \vdots&\vdots&\vdots&\vdots \\X_n & Y_n & Z_n & 1\end{bmatrix}  \begin{bmatrix}a \\ b \\c \\ d \end{bmatrix} =0
	\label{eq:kinect1mirror}
\end{equation}
where $(X_i,Y_i,Z_i)$ is the 3D coordinates of the $i^{th}$ marker. A solution could be approximated by performing singular value decomposition (SVD) on the first matrix \cite{Szeliski2010}. Depth information estimated in practical environments, however, is usually affected by noise. The obtained mirror plane thus may have a significant deviation, especially when working on low-cost devices. Therefore, we applied a combination of RANSAC \cite{Fischler1981} and SVD to reduce the effect of outliers (noise) in order to get better results. The next section describes in detail the use of mirror surfaces in reconstructing a 3D point cloud.

\section{3D reconstruction}\label{sec:kinect1reconstruction}

According to our configuration, which consists of an object directly seen by a depth camera and 2 or more mirrors around, the object is represented in captured images as a collection of object's pieces including a real one and some virtual ones, i.e. behind mirrors. As mentioned, the whole object is formed by combining the real points directly seen in front of the camera with reflections of virtual pieces obtained via corresponding mirror planes.

In detail, given the internal matrix $K$ including camera parameters in Eq.~(\ref{eq:kinect1reproject}), a set of $n$ object pixels $\{\tilde{p}\}_n$ in the captured image and corresponding depth values $\{Z\}_n$, $m$ mirror planes $\{\pi\}_m$ and 2D object boundaries $\{\hat{b}\}_m$, our method for reconstructing a point cloud $\{P\}$ which represents the object is as follows:

\RestyleAlgo{boxruled}
\begin{algorithm}[!htb]%[ht]
%\SetAlgoLined
\SetKwFunction{Reproject}{Reproject}
\SetKwFunction{Reflect}{Reflect}
\KwData{$K,\{\tilde{p}\}_n, \{Z\}_n, \{\pi\}_m$, $\{\hat{b}\}_m$}
\KwResult{$\{P\}$}
\nl $\{P\} \leftarrow \emptyset$\;
 \For{$i\leftarrow 1$ \KwTo $n$}{
\nl			$P_i \leftarrow$ \Reproject{$\tilde{p}_i,Z_i,K$}\;
  \For{$j\leftarrow 1$ \KwTo $m$}{
		\If{$\tilde{p}_i$ inside $\hat{b}_j$ {\bf and} $P_i$ behind $\pi_j$}{
\nl			$P_i \leftarrow$ \Reflect{$P_i,\pi_j$}\;
			break\;
	 }
	}
\nl  $\{P\} \leftarrow \{P\}\cup P_i$\;
 }
 \caption{Reconstructing a raw point cloud of the object from a depth image.}
\label{algo:kinect1raw}
\end{algorithm}

In the Algorithm~\ref{algo:kinect1raw}, the reprojection at line 2 is performed based on Eq.~(\ref{eq:kinect1reproject}), and the reflection at line 3 is done according to the following equation \cite{Coxeter1967}
\begin{equation}
	P_r = P - 2 \| \hat{n} \| ^{-1}(P^\top\hat{n}+d)\hat{n}
	\label{eq:kinect1reflect}
\end{equation}
where $P_r$ is the point reflected from $P$ via a plane of parameters $(a,b,c,d)$, and $\hat{n}$ is the corresponding normal vector, i.e. $\hat{n}=[a,b,c]^\top$.

\begin{algorithm}[h]%[!htb]%[ht]
%\SetAlgoLined
\SetKwFunction{Reproject}{Reproject}
\SetKwFunction{Project}{Project}
\SetKwFunction{Abs}{Abs}
\SetKwFunction{Reflect}{Reflect}
\KwData{$V_{init}, K,\{\tilde{p}\}_n, \{Z\}_n, \{\pi\}_m$, $\{\hat{b}\}_m$}
\KwResult{$V_{carved}$}
$V_{carved} \leftarrow V_{init}$\;
$th \leftarrow t_0$\;
\ForEach{voxel $v \in V_{carved}$}{
	$\tilde{p} \leftarrow$ \Project{$v,K$}\;
	\If{$\tilde{p}\notin\{\tilde{p}\}_n$}{
		$v \leftarrow false$\;
		continue\;
	}
	$P \leftarrow$ \Reproject{$\tilde{p},Z_p,K$}\;
	$v \leftarrow true$\;
	\eIf{$\Abs(\|v\|-\|P\|)<th$}{
		continue\;}{
		\For{$j \leftarrow 1$ \KwTo m}{
			$v_j\leftarrow$\Reflect{$v,\pi_j$}\;
			$\tilde{p}_j \leftarrow$ \Project{$v_j,K$}\;
			\If{$\tilde{p}_j\notin\{\tilde{p}\}_n$ {\bf or} $\tilde{p}_j$ not inside $\hat{b}_j$}{
				$v \leftarrow false$\;
				break\;
			}
			$P_j \leftarrow$ \Reproject{$\tilde{p}_j,Z_{\tilde{p}_j},K$}\;
			\If{$\Abs(\|v_j\|-\|P_j\|)\geq th$}{
				$v \leftarrow false$\;
				break\;
			}
		}
	}
}
\caption{Reconstructing a volume of voxels representing the object, in which the assignment of Boolean values $true$ or $false$ to each voxel indicates that this voxel is kept or removed, respectively.$\newline$\textbf{Notation:} $\newline th$: a threshold related to the thickness of the reconstructed object boundary$\newline Z_{\tilde{p}}$: measured depth value at pixel $\tilde{p}$}
\label{algo:kinect1carving}
\end{algorithm}

When working on moving objects, the Algorithm~\ref{algo:kinect1raw} is an appropriate choice because it can run in real-time with a low computational cost. However, in some situations, one could want to reconstruct an object point cloud with a higher density. The space carving technique \cite{Kutulakos2000} is a suitable approach in these cases, especially with static objects. The overall idea of our workflow could be summarized in Algorithm~\ref{algo:kinect1carving} by following steps performed on each voxel of a predefined volume. In detail, we first compute the projected pixel based on the voxel coordinates and the calibrated internal camera matrix. The corresponding estimated depth $\|P\|$ is then compared with the voxel's depth $\|v\|$, and a deviation is calculated. The voxel is kept in the volume if such deviation is less than a predefined threshold, i.e. this voxel corresponds to a real point captured directly by the depth camera. Otherwise, the voxel is reflected through each mirror and the mentioned checking is repeated on each virtual reflection result. The voxel is removed from the volume if the deviation condition is not satisfied with at least one mirror. Beside the space carving approach presented here, a high-density cloud could be obtained by using an additional high-resolution camera and employing a registration between its images and captured depth frames. This method is not described in this paper since one objective of our work is to reduce the device price.

Given a voxel volume $V_{init}$ in front of the mirrors and input terms similar to the Algorithm~\ref{algo:kinect1raw}, the space carving is applied to create the corresponding object volume $V_{carved}$ as in Algorithm~\ref{algo:kinect1carving}. Let us notice that the origin of the coordinate system in this algorithm is the camera center. With another 3D space, a rigid transformation \cite{Legarda2004} between it and the camera space is required, and vector terms in the Algorithm~\ref{algo:kinect1carving} (e.g. voxel $v$, point $P$) thus need to be recalculated with respect to the camera center.

In some cases, the collection of object pixels $\{\tilde{p}\}_n$ may be defined as a group of points representing a region which contains the object instead of a set of true pixels. Depending on each application as well as visual properties of the object of interest, some additional conditions could be integrated into the two algorithms to reduce noise, i.e. reconstructed points which are not object's parts. In our experiments, our system employed such constrains including background subtraction and color filtering. This content is not described in this paper since it does not play a principal role in our proposed algorithms.

\section{Experiment}\label{sec:kinect1testing}
\subsection{Configuration and error measurement}
In order to evaluate our approach in reconstructing 3D object point clouds, we built a configuration of a depth camera and two mirrors. The camera employed in our experiments is a Microsoft Kinect 1, which provides depth information by emitting an IR dot pattern and matching it with the corresponding captured IR image. This device was selected because of its cheap price and good SDK with many functionalities \cite{Webb2012}. There were two objects used in our experiments consisting of a cylinder and a sphere. Reconstruction accuracy was estimated by fitting each resulting point cloud according to its true shape and then calculating an error based on the cloud and fitted geometric parameters. Root mean square error (RMSE) was determined according to fitted center and radius in the case of a sphere, and the main axis and radius for the cylinder. The mean value of such deviations was also estimated in order to provide another error type which is easy to visualize.

\subsection{Test on sphere}
\begin{figure}[h]%[ht]
	\centering
	\begin{picture}(240,330)
	\put(-8,145){\includegraphics[scale=0.72]{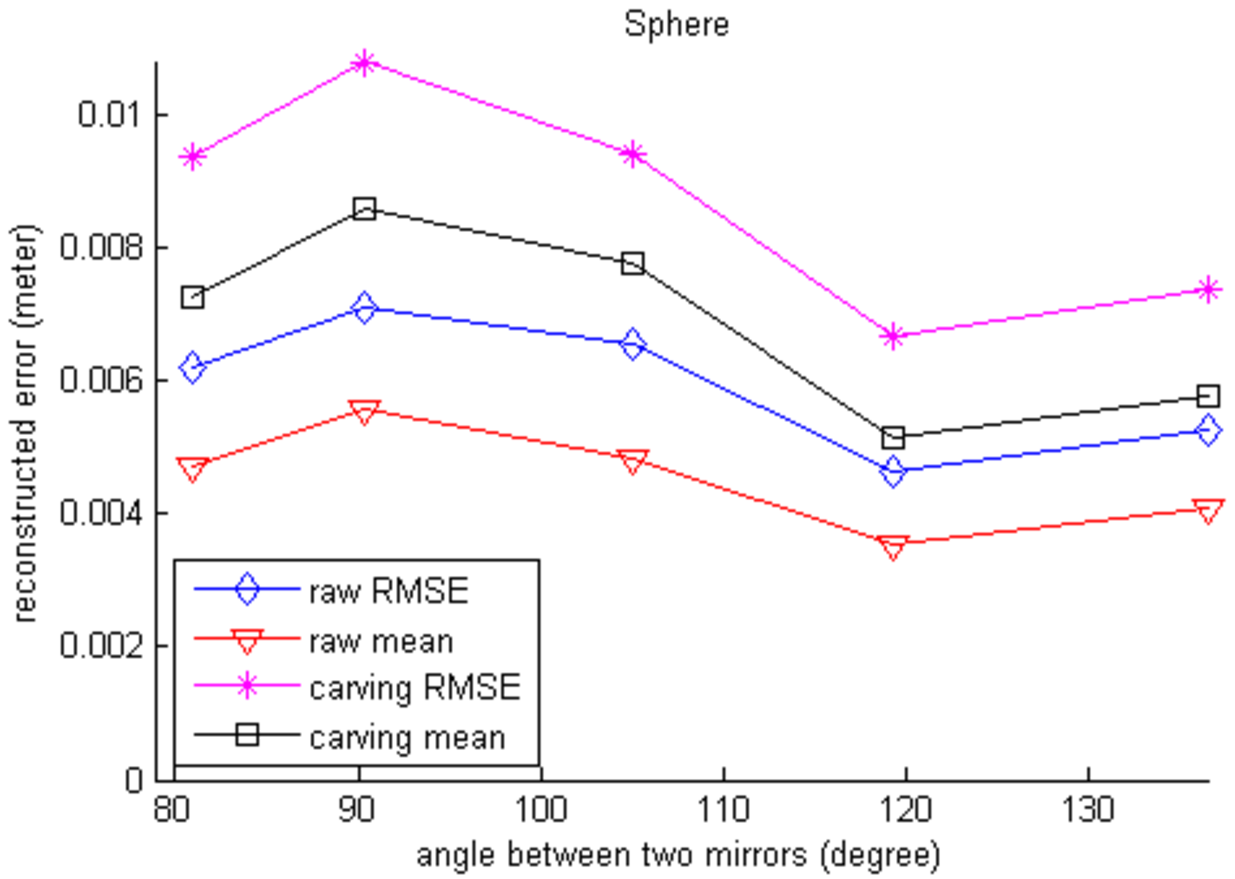}}
	\put(220,190){(a)}
	\put(0,0){\includegraphics[scale=0.55]{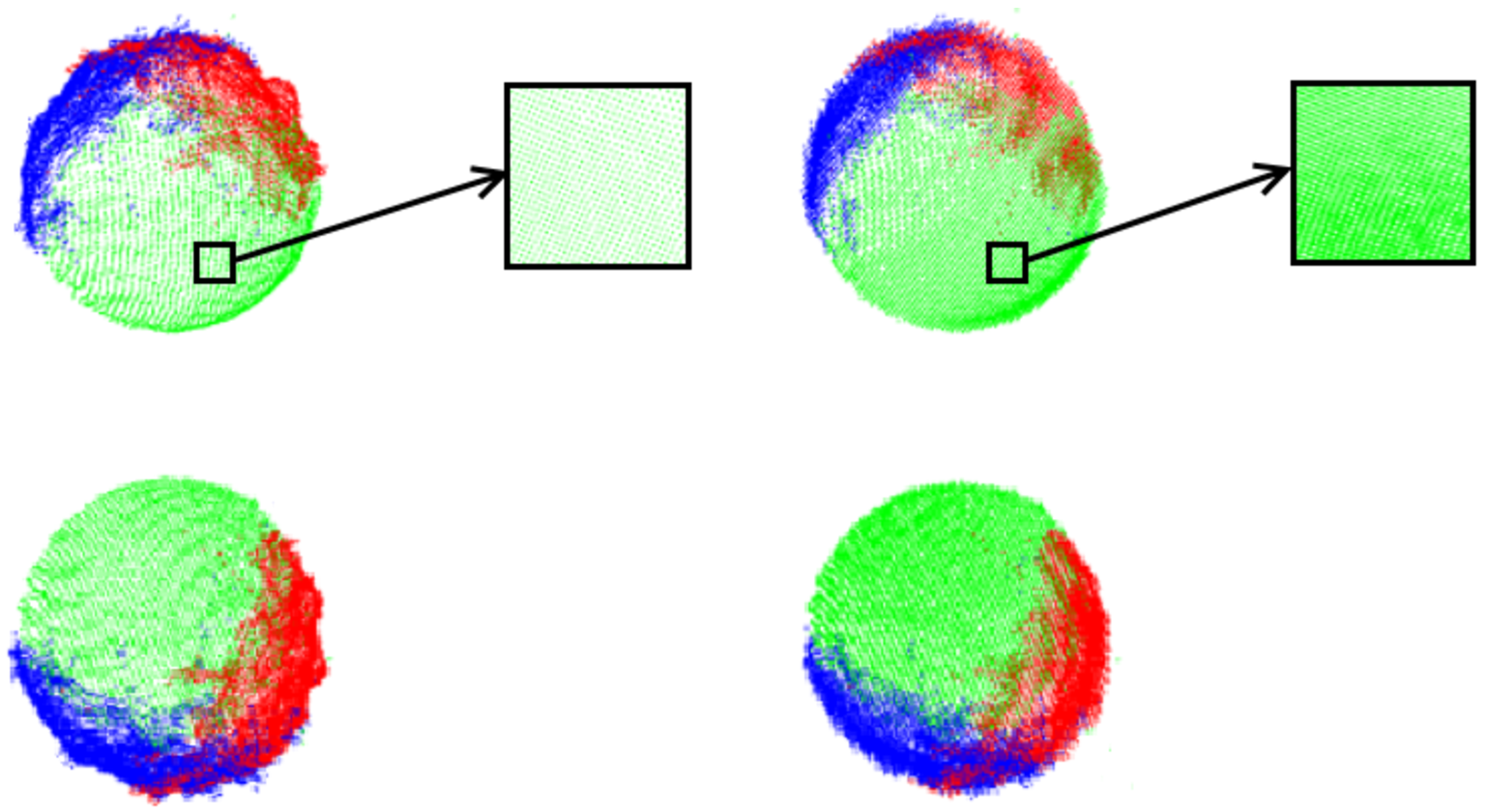}}
	\put(220,25){(b)}
	\put(60,64){raw cloud}
	\put(58,67){\line(-3,2){21}}
	\put(58,67){\line(-3,-2){21}}
	\put(189,64){carved cloud}
	\put(187,67){\line(-3,2){21}}
	\put(187,67){\line(-3,-2){21}}
	\end{picture}
	\caption{(a) Estimated fitting errors when reconstructing a sphere with different angles between the two mirrors, and (b) reconstructed point clouds which are seen at different viewpoints. The two terms ``raw'' and ``carved'' indicate the two point clouds reconstructed by Algorithm~\ref{algo:kinect1raw} and~\ref{algo:kinect1carving}, respectively. Different colors in cloud indicate points obtained from different sources, i.e. a depth camera and two mirrors.}
	\label{fig:kinect1sphere}
\end{figure}
With the spherical object, we performed the reconstruction at different angles between the two mirrors. The object shape was fitted by applying the RANSAC technique on the obtained cloud. The RMSE error was then estimated based on the equation
\begin{equation}
	\epsilon_{sphere} = \sqrt{ \frac{ \sum_{i=1}^n [dist(P_i, \tilde{c})- \Re]^2 }{n} } 
	\label{eq:kinect1sphere}
\end{equation}
where $dist$ is a function measuring Euclidean distance between two input coordinates, $P_i$ is the $i^{th}$ element of $n$ 3D points, $\tilde{c}$ and $\Re$ are the fitted sphere's center and radius, respectively. A simple mean error was also calculated as average of deviation values, i.e. $dist(P_i, \tilde{c})- \Re$. Our experimental results corresponding to the test on the sphere are shown in Fig.~\ref{fig:kinect1sphere}.

Both measured errors were less than 1 centimeter. The average length of estimated radii was 117 millimeters while the true value, which was manually measured, was 115 millimeters. The errors corresponding to the space carving approach were always greater than the other because of its higher cloud density and thicker surface. According to all four curves in Fig.~\ref{fig:kinect1sphere}, reconstruction errors tend to be lowest at a specific degree between mirrors (about $120^{\circ}$ in our experiment). We can thus expect that in an arbitrary configuration (in terms of distance between object and camera and/or mirrors) with two mirrors, there exists an angle between them which provides reconstructed object point clouds with lowest errors. This value can be estimated by trial-and-error.

\subsection{Test on cylinder}
\begin{figure}[h]
	\centering
	\begin{picture}(240,360)
	\put(-20,165){\includegraphics[scale=0.75]{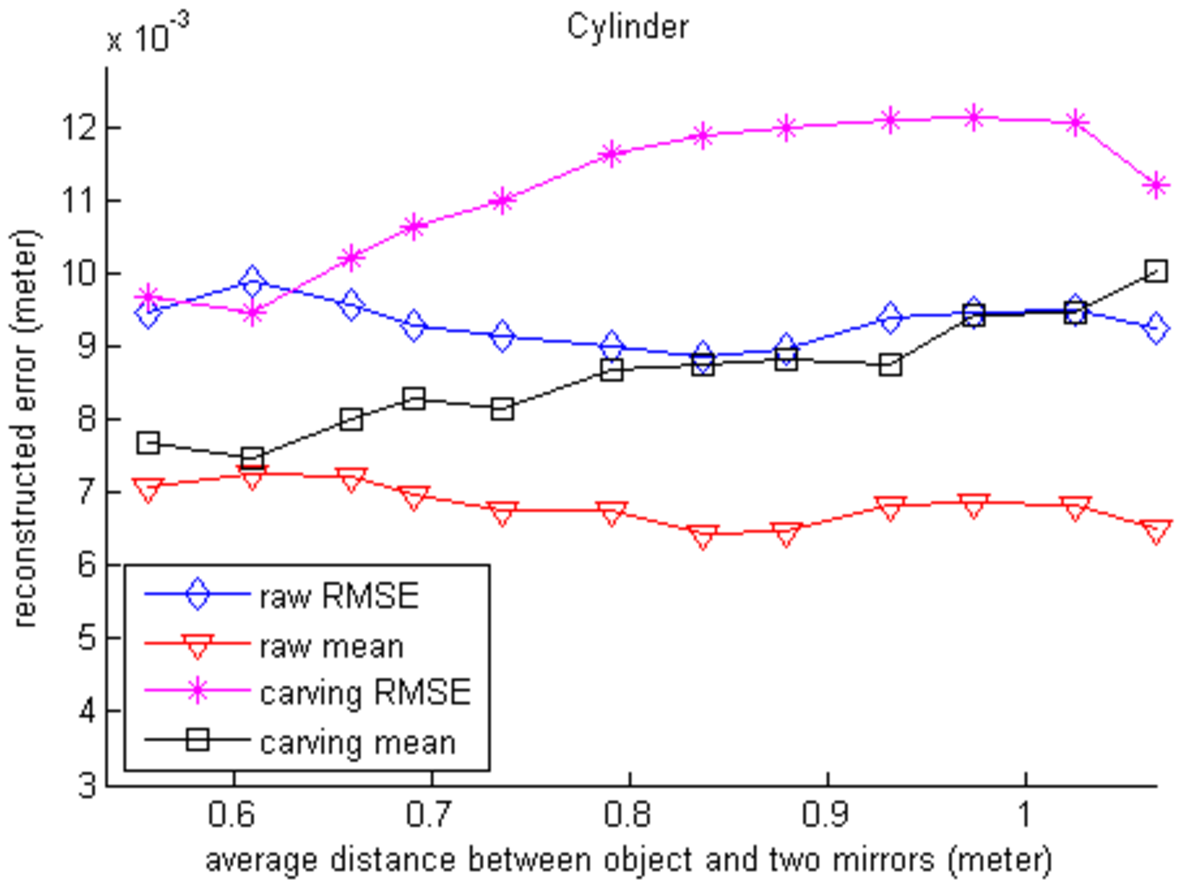}}
	\put(120,220){(a)}
	\put(0,0){\includegraphics[scale=0.4]{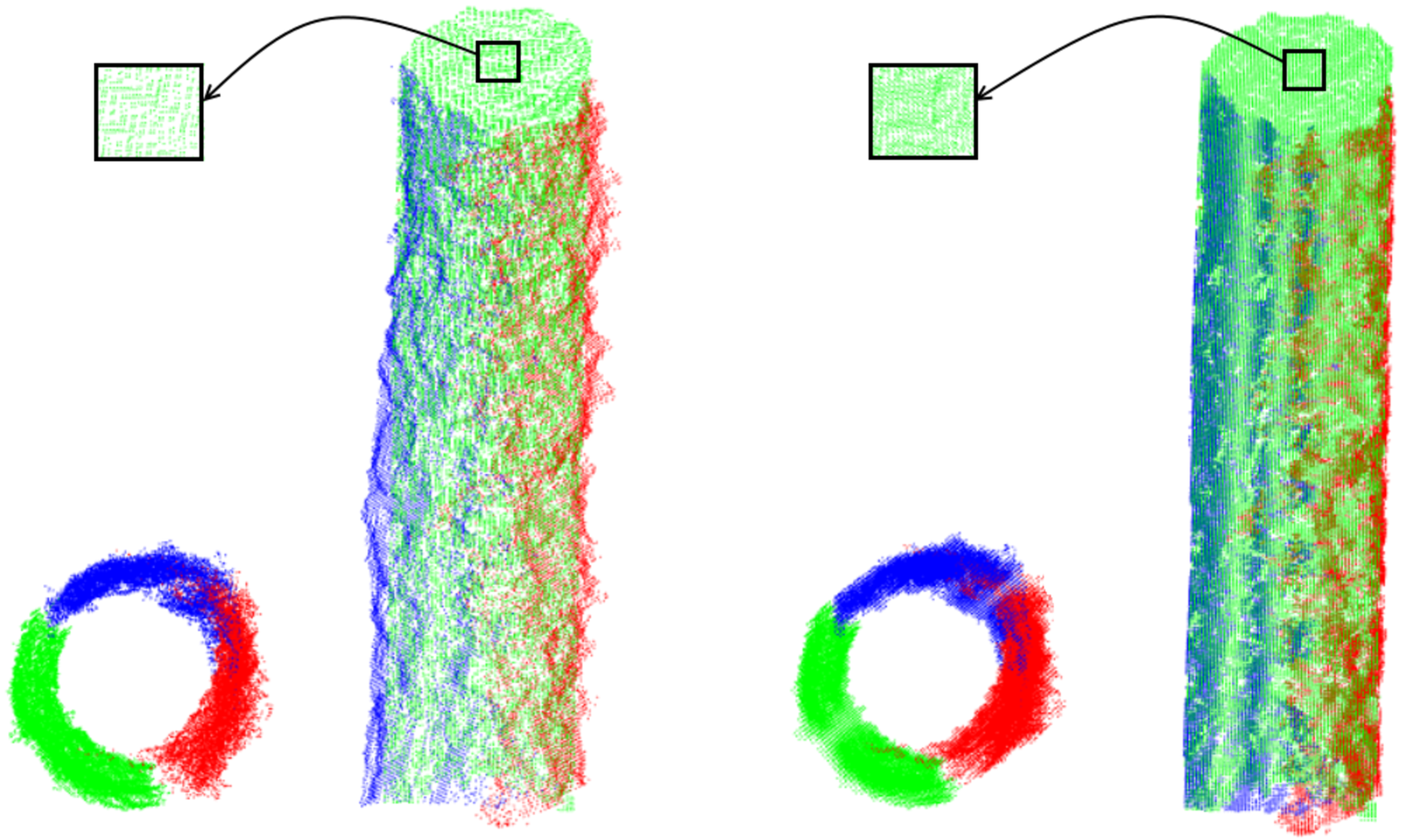}}
	\put(120,120){(b)}
	\put(0,80){\minibox{\hspace{0.04cm}\footnotesize{cloud obtained}\\\footnotesize{by Algorithm 1}}}
	\put(140,80){\minibox{\hspace{0.04cm}\footnotesize{cloud obtained}\\\footnotesize{by Algorithm 2}}}
	\end{picture}
	\caption{(a) Fitting errors when applying our approach on a cylinder at different (average) distances from the two mirrors, and (b) visualization of the clouds (top and side viewpoints), in which three colors correspond to points generated from the depth camera and 2 mirrors.}
	\label{fig:kinect1cylinder}
\end{figure}
When working on the sphere, we focused on its center coordinates and radius. With cylinder, the error measurement was performed based on the line equation of its axis and radius length. The experiment was done under different average distances between the object and the two mirrors. RANSAC was also employed for fitting the point cloud. The error was measured as
\begin{equation}
	\epsilon_{cylinder} = \sqrt{ \frac{ \sum_{i=1}^n [dist(P_i, \ell)- \Re]^2 }{n} } 
	\label{eq:kinect1cylinder}
\end{equation}
where $\ell$ is the straight line corresponding to the cylinder axis, $dist$ calculates the distance from a 3D point to a line, and $\Re$ is the fitted cylinder radius. The mentioned mean error is estimated by computing mean of deviations $dist(P_i, \ell)-\Re$. The obtained results are presented in Fig.~\ref{fig:kinect1cylinder} together with visualization of a pair of reconstructed clouds for top and side viewpoints. The true radius was 150 millimeters.

Similarly to our previous experiments, errors measured on raw clouds are less than on space carving results. These charts also show that when the distance between the tested object and mirrors increases, fitting errors of the former approach tend to slightly decrease while the latter one go in the opposite direction. This property can be explained based on captured 2D depth images. As usual, we can guess that depth information, which is directly measured, is usually more reliable than the reflected one. When increasing the mentioned distance, the number of pixels representing the object's part directly seen by the camera is also larger, and such quantity in mirror regions is reduced. Reconstruction errors thus become lower because of the increased proportion of more reliable information. This change, however, reduces the accuracy of the space carving approach because mirrors generate real 3D clouds in which points are more sparse. Sub-volumes of such regions thus could be carved wrongly producing larger errors. This drawback can be overcome by performing an interpolation on depth images to provide a sub-pixel level for 2D projection from each voxel. According to these properties, we can obtain good results when using either of both proposed algorithms by creating a configuration in which all devices are near each other around the object. In our experiments, the distance between the Kinect and tested objects was about 2 meters.

\begin{figure*}[h]
	\centering
	\includegraphics[]{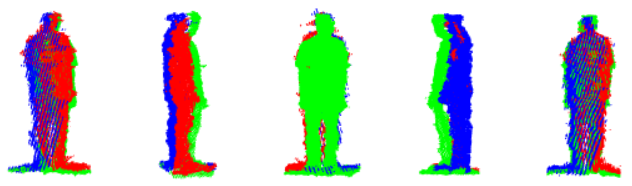}%scale=0.65
	\caption{Reconstructed point clouds of a human body with the same posture. This process was done by applying the proposed Algorithm 1 on noisy depth information. The points corresponding to ground can be easily removed as a post-processing step based on the ground calibration.}
	\label{fig:kinect1human}
\end{figure*}

As an illustration of a practical application, we also tried to reconstruct a 3D point cloud representing a human body based on the experimental configuration. The obtained results are shown in Fig.~\ref{fig:kinect1human}. We believe that these clouds are acceptable for realistic applications such as human gait or shape analysis.

\subsection{Implementation}
Our experimental system was executed on a medium-strength computer based on non-optimized C++ code and the two popular open source
libraries Point Cloud Library \cite{PCL2011} and OpenCV \cite{OpenCV}. Depth images in our work were captured with the largest possible resolution ($640 \times 480$ pixels) by the SDK version 1.8. Our system could be expected to reconstruct object point cloud using the Algorithm~\ref{algo:kinect1raw} in real-time since our non-optimized code processed each frame in about 0.2 seconds. The execution speed can even be increased by optimizing the source code of memory allocation and management as well as employing the power of parallel processing and/or multi-threading. Our approach could thus be integrated into vision-based systems without affecting significantly computational time.

\section{Conclusion}\label{sec:kinect1conclusion}
Throughout this paper, an approach which overcomes problems of synchronization has been proposed for reconstructing a 3D object point cloud. Our system can run with a low computational cost with low-cost devices since the proposed configuration employs only a matching-based depth camera together with a few mirrors. The two described algorithms, i.e. combination of reflected points and space carving, are appropriate for working on dynamic (e.g. a walking person on a treadmill for health analysis) as well as static objects, respectively. In summary, our approach can play a significant role in a low-price 3D reconstruction system and can provide acceptable intermediate object models for a wide variety of practical applications in many research fields. In future work, we intend to integrate our method into problems of human gait analysis for health assessment.

\subsubsection*{Acknowledgments}
The authors would like to thank professor S\'{e}bastien Roy, 3D Vision Lab, DIRO, for his useful comments on this work. We also would like to thank the NSERC (Natural Sciences and Engineering Research Council of Canada) for having supported this work (Discovery Grant RGPIN-2015-05671). This work was also supported by University of Science and Technology, The University of Danang, code number of Project: T2018-02-03.

\bibliography{references}
\bibliographystyle{iclr2019_conference}

\end{document}